%% file: main.tex
\documentclass{article} 
\usepackage{iclr2020_conference,times}

\input{math_commands.tex}

\usepackage{natbib}

\usepackage{hyperref}
\usepackage{url}
\usepackage{graphicx}
\usepackage[ruled,vlined]{algorithm2e}
\usepackage{multirow}

\usepackage{mwe,tikz}\usepackage[percent]{overpic}

\title{Time adaptive reinforcement learning}

\author{Chris Reinke \\
Inria Grenoble Rh\^one-Alpes\\
France\\
\texttt{chris.reinke@inria.fr} \\
}

\newcommand{\N}{\mathbb{N}}

\iclrfinalcopy 
\begin{document}

\maketitle

\begin{abstract}
Reinforcement learning (RL) allows to solve complex tasks such as Go often with a stronger performance than humans. However, the learned behaviors are usually fixed to specific tasks and unable to adapt to different contexts. Here we consider the case of adapting RL agents to different time restrictions, such as finishing a task with a given time limit that might change from one task execution to the next. We define such problems as \textit{Time Adaptive Markov Decision Processes} and introduce two model-free, value-based algorithms: the \textit{Independent $\gamma$-Ensemble} and the \textit{n-Step Ensemble}. In difference to classical approaches, they allow a zero-shot adaptation between different time restrictions. The proposed approaches represent general mechanisms to handle time adaptive tasks making them compatible with many existing RL methods, algorithms, and scenarios. 

\end{abstract}

\section{Introduction}

Modern reinforcement learning (RL) algorithms are able to learn impressive skills, such as playing Chess or Go, often with a higher competency than humans \citep{silver2017masteringchess, silver2017mastering}.
Nonetheless, the learned skills are highly specific.
Given a different context, such as to collect as many opponent pieces in the next 5 moves instead of winning the full game, the algorithms have to learn their behavior from scratch which is time consuming.
In difference, humans quickly adapt to such changes and still show a high performance.
The field of transfer learning \citep{taylor2009transfer, lazaric2012transfer} investigates how artificial agents can be more adaptive to such changes.
One of the areas where humans are very adaptive and which is the focus of this paper is in the presence of changing time restrictions for a task.
Take for example one of our daily routines, going out for lunch.
Several restaurants might exist in our neighborhood. 
Each requires a different amount of time to reach and provides food of different quality.
The general goal is to learn the shortest way to a good restaurant, but our specific objective might change from day to day.
One day we have a lot of time for lunch and we want to go to the best restaurant.
Another day we are under stress and we want to go to the best restaurant given a fixed time limit.


Standard model-free RL algorithms such as Q-learning \citep{watkins1989learning, watkins1992q} would learn such a task by defining a specific reward function for each objective.
For example, to learn the way to the best restaurant the reward function would simply return the food quality of a reached restaurant.
Thus, an agent will learn the shortest way to the best restaurant.
If the agent is under stress and has to judge between food quality and the invested time, the reward function could include a punishment (a negative reward) for each performed step until a restaurant is reached.
Thus, the agent might not go to the best restaurant, but to one that is closer.
With this approach, each objective represents a different task for which a new policy has to be learned.
But learning takes time and an objective might change from one episode to another. 
Moreover, the number of possible objectives can be infinite.
As a result, classical agents might not have time to learn an appropriate policy for each objective.

We formalize this new type of adaptation scenario in form of \textit{Time Adaptive MDPs} which are extensions of standard MDPs \citep{sutton1998reinforcement}.
Moreover, we propose two modular, model-free algorithms, the \textit{Independent $\gamma$ - Ensemble} (IGE) and the \textit{n-Step Ensemble} (NSE), to solve such scenarios.
Both learn several behaviors by their modules in parallel, each for a different time-scale, resulting in a behavioral library.
Given a change in the objective the most appropriate behavior from the library can be selected without need for relearning.
Both algorithms are inspired by neuroscientific findings about human decision-making which suggests that humans learn behaviors not only for a specific time-scale, but for several time-scales in parallel \citep{tanaka2007serotonin}.

\section{Time Adaptive MDPs}

We formalize time adaptive reinforcement learning tasks in form of Time Adaptive MDPs:
\begin{equation*}
	\mathrm{TA-MDP} \left( A, S, P, R, J \right) ~.
\end{equation*}
TA-MDPs are extensions of standard MDPs by having a finite or infinite set of objective functions $J = (f_1, f_2, \ldots)$ with $f_k: \R \times \N \mapsto \R$.
Each objective function $f_k(R, T)$ evaluates the agent's performance in regard to its total collected reward $R = \sum_{t=1}^T r_t$ and the number of time steps $T$ until it reached a terminal state $S^G \subset S$ for a single episode.
During each episode, one objective is active and the goal is to maximize the expectation over it while using a minimum number of steps:
\begin{equation}
\label{eq:tamdp_goal}
	\min_{\E[T]} \max_{\pi_k} \E_{\pi_k} \left[ f_k(R, T) \right] ~,
\end{equation}
where $\pi_k$ is the policy used for objective $f_k$.
Objective functions are monotonic increasing with reward $R$ and decreasing with time $T$, i.e.\ getting more reward and using less time is better.
The agent knows which objective function is active and its mathematical expression. 
The agents goal is to learn a policy to optimize this objective. 

Our restaurant selection example can be formalized as a TA-MDP.
The agent's location defines the state space $S$ and its actions $A$ are different movement directions.
Restaurants are terminal states $s \in S^G$.
The reward function $R(s)$ indicates the food quality of a restaurant after reaching it.
The two time restrictions are represented as objective functions: $J = (f_1, f_2)$.
The first objective is to go to the restaurant with highest food quality: $f_1(R,T) = R$.
The second objective sets a strict time limit of five steps and gives a large punishment if more steps are needed:
$f_2(R,T) = R$ if $T \leq 5$ otherwise $-10$.
Depending on the active objective, the optimal policy, i.e.\ where to eat, changes.

The major challenge of TA-MDPs is that the number of objective functions can be infinite and that each episode can have a different objective.
Thus, an agent might experience a certain objective only once.
As a result, it needs to immediately adapt to it.

\section{Algorithms}

We propose two algorithms to solve TA-MDPs.
Both are modular and learn a set of policies $\Pi = (\pi_1, ..., \pi_M)$.
The policies represent optimal behaviors on different time scales.
Given the active objective $f_k \in J$ of the episode, one of the policies is selected at the start of the episode and used throughout it.
The goal is to select the most appropriate policy from the set $\Pi$.
To accomplish this, both algorithms learn for each of its policies $\pi_i$ the expected total return $R_i(s_t) =  \E_{\pi_i,s_t} \left[ \sum_{k=1}^{T-t} r_{t+k} \right]$ 
and the expected number of steps until a terminal state is reached
$T_i(s_t) = \E_{\pi_i,s_t} \left[ T \right]$.
Based on the expectations the policy $\pi_j$ is selected at the start of the episode ($t=0$) which maximizes the active objective $f_k$ while minimizing the number of steps (Eq.~\ref{eq:tamdp_goal}):
\begin{equation}
    \label{eq:opt_module}
	j = \argmin_{T_i} \argmax_{\pi_i \in \Pi} f_k ( R_i(s_0), T_i(s_0))  ~. 
\end{equation}
An important restriction of the method is that the selection of the policy is dependent on an approximation of the expected outcome for the objective (Eq. \ref{eq:tamdp_goal}) by using the expectations over the total return $R(s)$ and number of steps $T(s)$ as input to the objective function:
\begin{equation*}
    \begin{array}{rcl}
         \E_{\pi_i}\left[ f_k(R(s),T(s)) \right] & \approx & f_k(R_i(s), T_i(s))\\
         & & =  f_k(\E_{\pi_i}[R(s)], \E_{\pi_i}[T(s)]) ~.
    \end{array}
\end{equation*}
This approximation is not correct for all types of objective functions but provides often a good heuristic to select an appropriate policy.
The proposed algorithms are introduced in the next sections.
They differ in their way to learn policy set $\Pi$.




\subsection{The Independent $\gamma$-Ensemble (IGE)}

The Independent $\gamma$-Ensemble (IGE) is composed of several modules (Alg.~\ref{alg:ga_ige_prediction}). 
The modules are independent Q-functions with different discount factors: $\Gamma = (\gamma_1, \ldots, \gamma_M)$ with $\gamma_i \in (0,1)$, similar to the Horde architecture \citep{modayil2014multi}:
\begin{equation}
\label{eqn:q_gamma_function}
\begin{array}{lll} 
	\textrm{for each }\gamma \in \Gamma:~~~
	Q_{\gamma}(s_t,a_t)
	& = &\E\big[ r_{t+1} + \gamma r_{t+2} + \gamma^2 r_{t+3} + \cdots \big] \\
	~& = & \E\big[ r_{t+1} + \gamma \max_{a_{t+1}} Q_\gamma(s_{t+1},a_{t+1}) \big]~.
\end{array}
\end{equation}
The factor $\gamma$ defines how strong future reward is discounted.
For low $\gamma$'s the discounting is strong and the optimal behavior is to maximize rewards that can be reached on a short time scale.
Whereas, for high $\gamma$'s the discounting is weak, resulting in the maximization of rewards on longer time scales.
As a result, each Q-function defines a different policy $\pi_{\gamma_i}$ and the IGE learns a set of policies via its modules: $\Pi_{\text{IGE}}=(\pi_{\gamma_1}, \ldots, \pi_{\gamma_M})$.
The values of each module are learned by Q-learning. 
Because Q-learning is off-policy the values of all modules can be updated by the same observations.

Additionally to the values, each module learns the expected total return $R_\gamma(s)$ and number of steps to reach a terminal state $T_\gamma(s)$ for its policy.
The expectations are used to select the appropriate policy at the beginning of an episode (Eq.~\ref{eq:opt_module}).
Both expectations can be incrementally formulated similar to the Q-function and also learned in a similar manner.
After an observation $(s_t,a_t,r_t,s_{t+1})$, the expectations of all modules which have action $a_t$ as greedy action ($\forall \gamma \in \Gamma: ~ a_t = \argmax_{a} Q_\gamma(s_t,a)$) are updated by:
\begin{equation}
    \label{eq:ige_r}
         R_\gamma(s_t) \leftarrow R_\gamma(s_t) + \alpha \left(r_t + R_\gamma(s_{t+1}) - R_\gamma(s_t)\right) ~,  \\
\end{equation}
\begin{equation}
    \label{eq:ige_t}
         T_\gamma(s_t) \leftarrow T_\gamma(s_t) + \alpha \left(1 + T_\gamma(s_{t+1}) - T_\gamma(s_t)\right) ~, 
\end{equation}
where $\alpha$ is the learning rate parameter also used for updates of Q-values.

\begin{figure}[b!]
\centering

\setlength\tabcolsep{0pt}
\begin{tabular}{@{}ccc@{}}

(a) Example MDP & (b) Choices & (c) $\gamma$-Curve \\

 \includegraphics[width=0.32\linewidth]{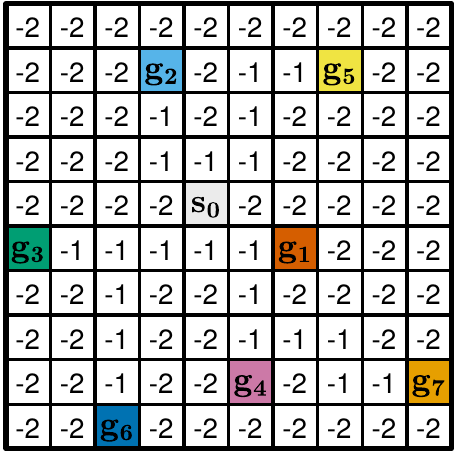}
 ~
 &
 \includegraphics[width=0.32\linewidth]{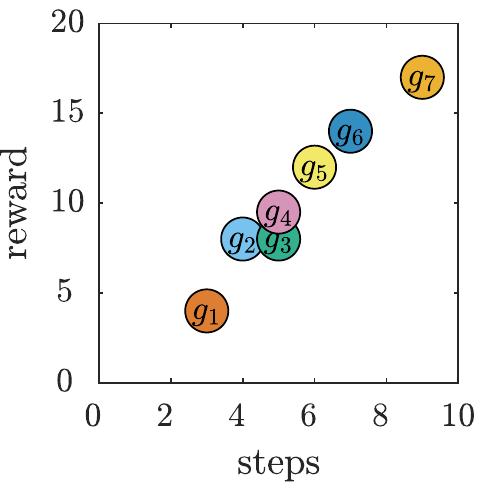}
 &
 \includegraphics[width=0.32\linewidth]{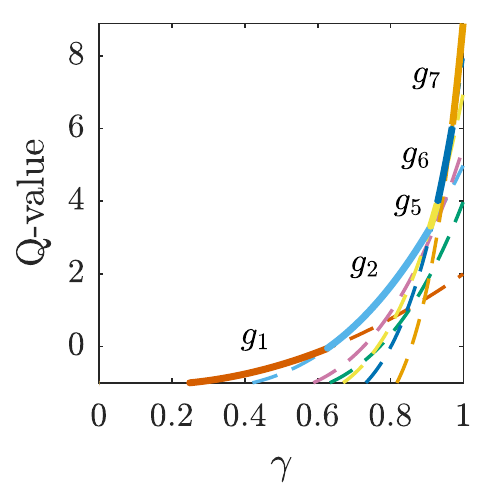}
 
\end{tabular}

 \caption [OA-MDP for evaluating objective-adaptive algorithms in general tasks]
 {
 	(a) 2D grid-world with 7 terminal states ($g_1,\ldots,g_7$) and one start state $s_0$.
 	The agent has 4 actions (move north, east, south, west) and moves with $\Pr = 0.1$ in a random direction. 
 	Transitions result in negative reward until a terminal state is reached. 
 	(b) If the agent starts in $s_0$ it has several choices each represented as the expected reward $R$ and number steps $T$ of the optimal policies to reach each terminal state.
 	All choices except going to $g_3$ are part of the Pareto optimal choice set.
 	(c) Discounted values $V_\gamma(s_0,g)$ for each choice (broken lines). 
 	The solid line represents the values the IGE learns ($\max_g V_\gamma(s_0,g)$).
 	The IGE does not learn policies to go to $g_3$ and $g_4$.
 }
 \label{fig:mdp}
\end{figure}

The IGE has the restriction that it does not guarantee to learn the Pareto optimal set of policies in regard to the expected total reward $R$ and number of steps $T$.
The MDP in Fig.~\ref{fig:mdp} shows such a case.



\subsection{The n-Step Ensemble (NSE)}

We propose a second algorithm (Alg.~\ref{alg:nse}), the n-Step Ensemble (NSE), to overcome the restriction of the IGE.
It is able to learn the set of Pareto optimal policies.
Similar to the IGE, the NSE also consists of several modules.
Each module is responsible to optimize the expected total reward for $n$ number of steps into the future with $N = (1, \ldots, M)$.
Each module learns a value function representing the optimal total reward that can be reached in $n$ steps:
\begin{equation}
\label{eq:nse_q}
    \textrm{for each } n \in N: ~~~ Q_n(s_t,a_t) = \E\left[r_{t+1} + r_{t+2} + \ldots + r_{t+n} \right] ~.
\end{equation}
One extra condition is that a terminal state should be reached within $n$ steps.
If this is not possible, then the policy is learned which reaches a terminal state with a minimal number of steps.
Similar to the standard Q-function the values can be incrementally defined by the sum of the immediate expected reward $r_{t+1}$ and the Q-value for the optimal action $a^*_{n-1}$ of the next state $s_{t+1}$.
In difference, the Q-value for the next state is from the module responsible to optimize the total reward for $n-1$ steps \citep{harada1997time}:
\begin{equation}
\label{eq:nse__q_iterative}
	\begin{array}{lcl}
	Q_1(s_t,a_t) &=& \E\left[r_{t+1}\right] \\	 	
	Q_n(s_t,a_t) &=& \E\left[r_{t+1} + Q_{n-1}(s_{t+1},a^*_{n-1}(s_{t+1})) \right] ~. \\
	\end{array}
\end{equation}
The learning of the Q-values $Q_n$, expected reward $R_n$, and number of steps $T_n$ for each module is done by Q-learning similar to the IGE. 
After a transition is observed all modules $N =(1,\ldots, M)$  are updated in parallel.


\section{Experimental evaluation}

The IGE and NSE were compared to classical Q-learning in a stochastic grid-world environment (Fig.~\ref{fig:mdp}).
It consists of 7 terminal states.
For each terminal state exists an optimal path from the start state for which the agent receives a punishment of $-1$ per step (otherwise $-2$).
Reaching a terminal state results in a positive reward where more distant goals result in a higher reward.
Agents had to adapt to 9 different objectives $J = (f_1, \ldots, f_9)$ (see Fig.~\ref{fig:all_results} for their formulation).
For each objective $6000$ episodes were performed to evaluate how long the agents need to adapt before the task switched to the next objective.

The classical Q-learning algorithm learned for each objective an independent Q-function.
Its reward function was defined by the outcome of the active objective function $f_k$ after the agent reached a terminal state.
This formulation does not fulfill the Markov assumption because the reward for reaching a terminal state depends on the whole trajectory.
As a result, the MDP is partially observable for the agent.
To reduce this problem the current time step $t$ was used as an extra state dimension for the classical Q-learning agent, improving its performance.

\begin{figure}[t!]
\centering

\setlength\tabcolsep{1pt}
\begin{tabular}{@{}cccc@{}}
$f_1 = R$ & 
$f_7 = \left\lbrace \begin{array}{cl}R   & |~ T \leq 5\\ -10& |~ T > 5 \end{array} \right.$ & 
$f_8 = \frac{R}{T}$ \\ 

\raisebox{-.5\height}{\includegraphics[width=0.32\linewidth]{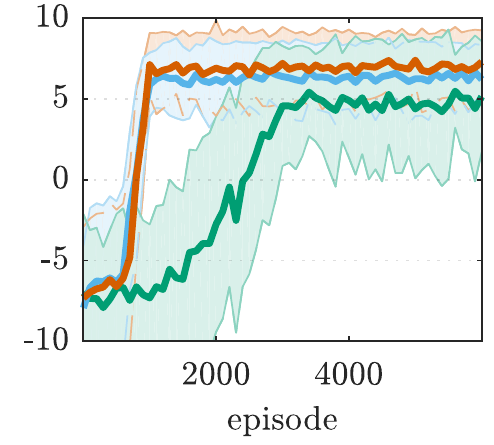}} & 
\raisebox{-.5\height}{\includegraphics[width=0.32\linewidth]{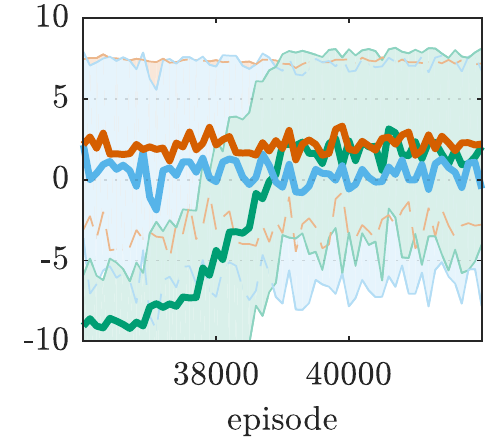}} & 
\raisebox{-.5\height}{  \begin{overpic}[width=0.32\linewidth]{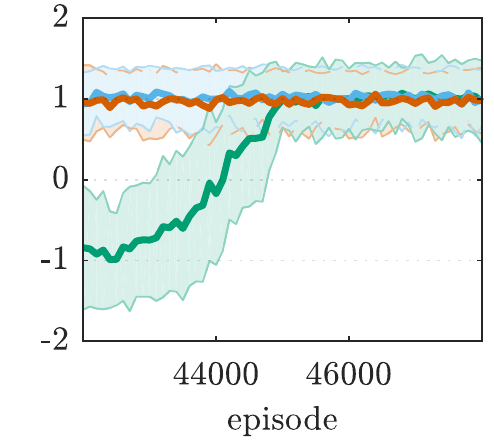}
                            \put(55.5,21){\includegraphics[width=0.135\linewidth]{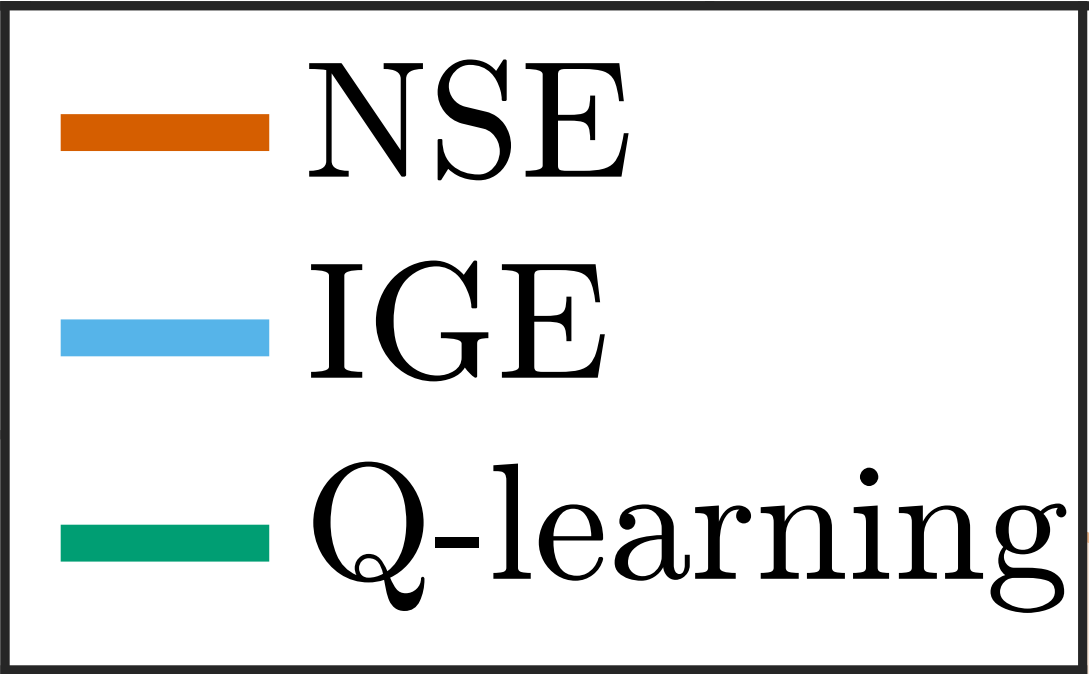}}  
                        \end{overpic}
                    } \\ 
\end{tabular}
\caption{
	Results for the outcome of the objective function $f_k(R,T)$ per episode show that the NSE and IGE adapt immediately to new objectives after the initial learning phase.
	3 of the 9 phases are shown where each phase has a different objective function (Fig.~\ref{fig:all_results} in the appendix lists all results).
	The plots show the mean and standard deviation over $100$ runs per algorithm.
 } 
 \label{fig:selected_results}
\end{figure}

The results show that the IGE and NSE outperformed classical Q-learning in terms of their adaptation to new objectives (Fig.~\ref{fig:selected_results} and \ref{fig:all_results}).
They were able to adapt immediatley to a new objective after they learned their set of policies during the initial phase.
Q-learning had to learn the task for each objective from scratch needing approximately 3000 episodes per objective.

\section{Conclusion}

We introduced Time Adaptive MDPs.
They confront RL agents with the problem of quickly adapting to changing objectives in terms of time restrictions.
Two algorithms are proposed (IGE and NSE) which learn a behavioral library of policies that are optimal on different time scales.
The agents can switch immediately between these policies to adapt to new and unseen objectives allowing zero-shot adaptation.
The NSE has the advantage over the IGE to learn the Pareto-optimal set of policies in terms of expected reward and time.
Nonetheless, the NSE is dependend on discrete time steps whereas the IGE can be used for continuous time MDPs \citep{doya2000reinforcement}.
Although we used for both algorithms tabular Q-learning to learn the values of their modules, the general scheme of the methods is independent of this choice.
The algorithms can also be combined with actor-critic or policy search algorithms and different function approximators such as deep networks.
This allows the methods to tackle various problem scenarios which we plan to show in future research.

\subsubsection*{Acknowledgments}
I want to thank Kenji Doya and Eiji Uchibe for their helpful supervision of this project.
Moreover, I want to thank Pierre-Yves Oudeyer and Cl´ement Moulin-Frier for their helpful comments.

\bibliography{iclr2020_conference}
\bibliographystyle{iclr2020_conference}

\newpage
\appendix

\section{Algorithmic Details}

\subsection{The Independent $\gamma$-Ensemble (IGE)}
\label{c:ige_framework}

\begin{algorithm}[b!]
\caption{Independent $\gamma$-Ensemble (IGE)}
\label{alg:ga_ige_prediction}

\SetAlgoLined
\DontPrintSemicolon
\SetKw{KwFrom}{from}
\KwIn{\\
~~Discount factors: $\Gamma = (\gamma_1, \ldots , \gamma_M)$ with $\gamma_i \in (0,1)$\\
~~Learning rate: $\alpha \in [0,1]$ \\
~~Exploration rate: $\epsilon \in [0,1]$}

\BlankLine

initialize $Q_\gamma(s,a)$, $R_\gamma(s)$, and $T_\gamma(s)$ to zero \\

\Repeat ({~(for each episode)}){termination}{
	initialize state $s$, and objective $f_k$\\
	
	\BlankLine
	
	\tcp{select as active module $\tilde{\gamma}$ a module that maximizes $f_k$}	
	
	$\tilde{\Gamma} \leftarrow \argmax_{\gamma} f_k(R_{\gamma}(s), T_{\gamma}(s))$ 	

    $\tilde{\gamma} \leftarrow \argmin_{\gamma \in \tilde{\Gamma}} T_{\gamma}(s)$

	\BlankLine

	\Repeat ({~(for each step in episode)}){$s$ is terminal-state}{

		\tcp{choose action $\epsilon$-Greedy}
		$a \leftarrow 	\begin{cases}  
						\argmax_a Q_{\tilde{\gamma}}(s,a) & | \text{ with } \Pr = 1-\epsilon \\
						\text{random action} & | \text{ with } \Pr = \epsilon					
						\end{cases}$
		
		$r$, $s' \leftarrow$ take action $a$, observe outcome
		
		\ForAll {$\gamma \in \Gamma$} {
		
			
			$\delta_Q = r + \gamma \max_{a'} Q_\gamma(s',a') - Q_\gamma(s,a)$
			
			$Q_\gamma(s,a) \leftarrow Q_\gamma(s,a) + \alpha \delta_Q$ 

			\BlankLine
			
			\tcp{update $R$ and $T$ only if the greedy action was used}
			
			\If {$a = \argmax_{\bar{a}} Q_\gamma(s,\bar{a})$} { 
				$R_\gamma(s) \leftarrow R_\gamma(s) + \alpha (r + R_\gamma(s') - R_\gamma(s))$
				
				$T_\gamma(s) \leftarrow T_\gamma(s) + \alpha (1 + T_\gamma(s') - T_\gamma(s))$
				
			}
			
		}
		$s \leftarrow s'$
	}
}
\end{algorithm}

The IGE (Alg.~\ref{alg:ga_ige_prediction}) learns a set of policies $\Pi_{\text{IGE}}=(\pi_{\gamma_1}, \ldots, \pi_{\gamma_M})$ via independent modules called $\gamma$-modules.
Each module is comprised of a Q-function (Eq.~\ref{eqn:q_gamma_function}) with a distinct discount factor $\gamma_i \in (0,1)$.
The number of modules $M$ and their discount parameters $\Gamma = (\gamma_1, \ldots, \gamma_M)$ are meta-parameters of the algorithm.
The Q-values of each module are learned by Q-learning, i.e.\ after an observation of $(s_t,a_t,r_t,s_{t+1})$ the values of all modules in $\Gamma$ are updated by:
\begin{equation}
    \label{eq:q_update}
    Q_\gamma(s_t,a_t) \leftarrow Q_\gamma(s_t,a_t) + \alpha (r_{t+1} + \max_a Q_\gamma(s_{t+1}, a) - Q_\gamma(s_t, a_t)) ~,
\end{equation}
where $\alpha \in (0,1)$ is the learning rate.
Because of Q-learning's off-policy nature, the values of all modules can be updated by the same observations.

Moreover, each module learns for its policy $\pi_i$ the expected total reward:
\begin{equation*}
\label{eq:expectation_r}
\begin{array}{rcl}
	R_i(s_t) & = &  \E_{\pi_i,s_t} \left[ \sum_{k=1}^{T-t} r_{t+k} \right] \\
		~		& = &  \E_{\pi_i,s_t} \left[ r_{t+1} \right] + \ldots + \E_{\pi_i,s_t} \left[ r_{T} \right]\\
		~		& = &  \E_{\pi_i,s_t} \left[ r_{t+1} \right] +  \E_{\pi_i,s_t} \left[ R_i(s_{t+1})\right] ~, 
\end{array}
\end{equation*}
and the expected number of steps until a terminal state is reached:
\begin{equation*}
\label{eq:expectation_t}
\begin{array}{rcl}
	T_i(s_t) 	& = &  E_{\pi_i,s_t} \left[ T \right] \\
		~ 			& = &  E_{\pi_i,s_t} \left[ \sum_{k=t}^{T} 1 \right] \\
		~			& = &  1 +  \Pr(t+2|\pi_i) + \cdots + \Pr(T|\pi_i)\\
		~			& = &  1 +  E_{\pi_i,s_t} \left[ T(s_{t+1})\right] ~.
\end{array}
\end{equation*}
Both are learned via a Robbins-Monro approach for stochastic approximation \citep{robbins1951stochastic} defined in Eq.~\ref{eq:ige_r} and \ref{eq:ige_t}.

The policy that the IGE follows is defined by one of its modules $\tilde{\gamma}$.
It is selected at the beginning of an episode with the goal to maximize the objective function according to Eq.~\ref{eq:tamdp_goal}.
The modules policy is then used for the action selection during the whole episode.
An $\epsilon$-Greedy approach is used for exploration.

The IGE has two restrictions.
First, it does not guarantee to learn the Pareto optimal set of policies.
Second, it can not handle episodic environments where the agent is able to collect positive rewards by circular trajectories that do no end in a terminal state.

The first restriction of the IGE is that it is not guaranteed to learn the Pareto optimal set regarding the expected reward and time.
The MDP used for the experiments shows such a case (Fig.~\ref{fig:mdp}, a).
The optimal trajectory to goal $g_4$ is part of the Pareto optimal set (Fig.~\ref{fig:mdp}, b), but it is not part of the IGE policy set (Fig.~\ref{fig:mdp}, c).
As a result, the IGE cannot find the optimal policy for objective $f_7$ that has $g_4$ as optimal solution.
Nonetheless, optimality can be guaranteed for a subset of goal formulations.
The IGE converges to the optimal policy for objectives that maximize the exponentially discounted reward sum $\E\left[\sum_{t=0}^{T-1} \gamma^t r_t \right]$.
Depending on the discount factor $\gamma$, the Q-values of the corresponding $\gamma$ module will converge to the optimal value function \citep{watkins1992q,tsitsiklis1994asynchronous}.
This includes also the case of maximizing the expected total reward sum $\E\left[\sum_{t=0}^{T-1} r_t \right]$, because this is the same objective as for $\gamma = 1$.
Most interestingly, it is possible to prove the convergence of the IGE upon the optimal policy for the average reward $\frac{R}{T}$ in MDPs which are deterministic and where non-negative reward is only given if a terminal state is reached \citep{reinke2017average}.
For other objectives, the IGE can be viewed as a heuristic that does not guarantee optimality, but that produces often good results with the ability to immediately adapt to a new objective.

\begin{figure}[b!]
	\centering
	
	a) Circular, positive reward MDP
	
	\vspace{0.1cm}
	\includegraphics{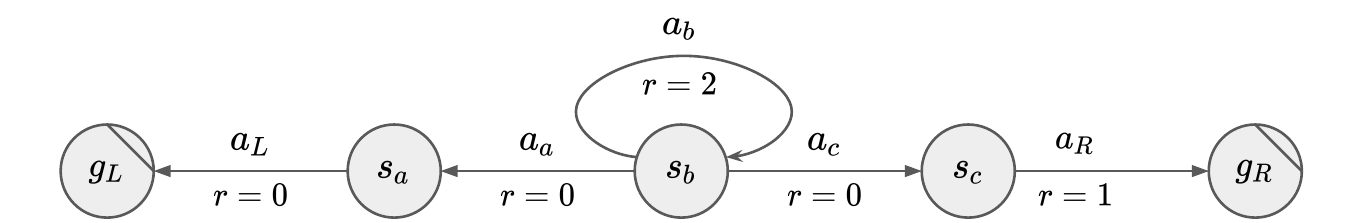} 
	\vspace{0.4cm}

    b) Q-Values and expected rewards and number of steps for each state and action
    
    \vspace{0.1cm}
	\begin{tabular}{cc}

		\multicolumn{2}{c}{
		
			\begin{tabular}{@{}|l|c|ccc|c|@{}}
    			\hline
    			\multirow{2}{*}{$Q_n$}		& $s_a$             &  \multicolumn{3}{c|}{$s_b$} 	& $s_c$ \\
    			\cline{2-6}
    										& $a_L$		        & $a_a$		& $a_b$		& $a_c$		& $a_R$\\
    			\hline
    			$Q_1$						& $0$	            & $0$       & $2$       & \textbf{0}       & $1$\\
    			$Q_2$						& $0$	            & $0$		& $2$		& \textbf{1}       & $1$\\
    			$Q_3$						& $0$	            & $0$		& \textbf{3}       & $1$       & $1$\\
    			$Q_4$						& $0$	            & $0$		& \textbf{5}       & $1$       & $1$\\
    			\hline
			\end{tabular}
			}
			\vspace{0.2cm}		
			\\

			\begin{tabular}{@{}|l|c|ccc|c|@{}}
				\hline
    			\multirow{2}{*}{$R_n$}		& $s_a$     &  \multicolumn{3}{c|}{$s_b$} 	& $s_c$ \\
    			\cline{2-6}
    										& $a_L$		& $a_a$		& $a_b$		& $a_c$		& $a_R$\\
				\hline
				$R_1$						& $0$		& $0$		& $3$       & \textbf{1}       & $1$\\
				$R_2$						& $0$		& $0$		& $3$       & \textbf{1}       & $1$\\
				$R_3$						& $0$		& $0$		& \textbf{3}       & $1$       & $1$\\
				$R_4$						& $0$		& $0$		& \textbf{5}       & $1$       & $1$\\
				\hline
			\end{tabular}		

		&
		
			\begin{tabular}{@{}|l|c|ccc|c|@{}}
				\hline
    			\multirow{2}{*}{$T_n$}		& $s_a$             &  \multicolumn{3}{c|}{$s_b$} 	& $s_c$ \\
    			\cline{2-6}
										    & $a_L$		& $a_a$		& $a_b$		& $a_c$		& $a_R$\\
				\hline
				$T_1$						& $1$		& $2$		& $3$       & \textbf{2}       & $1$\\
				$T_2$						& $1$		& $2$		& $3$       & \textbf{2}       & $1$\\
				$T_3$						& $1$		& $2$		& \textbf{3}       & $2$       & $1$\\
				$T_4$						& $1$		& $2$		& \textbf{4}       & $2$       & $1$\\
				\hline
			\end{tabular}
		
	 	\\

	\end{tabular}
	
	\caption
 	{
		MDP example for which the IGE is not able to learn policies that reach the terminal states ($g_L$, $g_R$).
		The agent starts in state $s_b$.
		It can either go left ($a_a$), right ($a_c$), or stay ($a_b$).
		Going left results in a final reward of 0, whereas going right results in 1.
		Both ways need 2 steps to reach their terminal state.
		For the number of steps $n=1$ and $n=2$ the action to stay ($a_B$) has the maximum Q-value.
		The NSE uses information from $T_n$ and $R_n$ to identify action $a_b$ as the greedy action.
 	} 
 	\label{fig:problem_mdp}
\end{figure}

A second restriction of the IGE exists in environments where cyclic trajectories maximize the discounted reward sum and which do not end in a terminal state.
Fig.~\ref{fig:problem_mdp} illustrates such a MDP. 
The agent starts in state $s_b$, and can stay in this state via action $a_b$ receiving a reward of $2$ for every step, or it can go 2 steps left to end in terminal state $g_L$ and receive no reward, or it can go 2 steps right to finish in terminal state $g_R$ receiving a reward of 1.
For this MDP the optimal policy for each $\gamma \in (0, 1]$ is to chose action $a_b$ and stay in state $s_b$.
Therefore, the IGE can not learn a policy to reach a terminal state.

\subsection{The n-Step Ensemble (NSE)}

We propose the n-Step Ensemble (NSE) (Algorithm~\ref{alg:nse}) to overcome the restrictions of the IGE.
It is able to learn the set of Pareto optimal policies, and to learn policies that reach terminal states in circular environments such as the MDP in Fig.~\ref{fig:problem_mdp}.

Similar to the IGE, the NSE consists of several modules.
Each module is responsible to optimize the expected total reward for $n$ number of steps into the future (Eq.~\ref{eq:nse_q} and \ref{eq:nse__q_iterative}) with $N = (1, \ldots, M)$.
To handle circular environments such as in Fig.~\ref{fig:problem_mdp} an extra condition is added.
The agent should also reach a terminal state within $n$ steps.
If this is not possible, then the policy is learned which reaches a terminal state with a minimal number of steps.
This is accomplished by defining the optimal action $a^*$ not simply as the action maximizing the Q-value of a state.
Instead, it is defined using information about the total reward $R_n$, and the number of steps $T_n$ until a terminal state is reached:
\begin{equation*}
\label{eq_nstep_r}
	\begin{array}{lcl}
	R_1(s_{t},a_{t}) &=& \E\left[r_{t+1} + R_1(s_{t+1}, a^*_1(s_{t+1})) \right] ~,  \\
	R_n(s_{t},a_{t}) &=& \E\left[r_{t+1} + R_{n-1}(s_{t+1}, a^*_{n-1}(s_{t+1})) \right] ~,  \\
	\end{array}
\end{equation*}
\begin{equation*}
\label{eq_nstep_t}
	\begin{array}{lcl}
	T_1(s_t,a_t) &=& \E\left[1 + T_1(s_{t+1}, a^*_1(s_{t+1})) \right] ~, \\
	T_n(s_t,a_t) &=& \E\left[1 + T_{n-1}(s_{t+1}, a^*_{n-1}(s_{t+1})) \right] ~. \\
	\end{array}
\end{equation*}
The state-values of $R_n$ and $T_n$ are defined by $R_n(s) = R_n(s,a^*_n(s))$, and $T_n(s) = T_n(s,a^*_n(s))$.

Based on $Q_n$, $R_n$, and $T_n$ the optimal action $a^*_n(s)$ is defined by:
\begin{equation}
\label{eq:nstep_opt_action}
	\begin{array}{lcl}
	a^*_n(s) &=& \argmax_{a \in A^T_n(s)} R_n(s,a) ~~, \text{with} \\ 
	A^T_n(s) &=& \argmin_{a \in A^Q_n(s)} T_n(s,a) ~~, \text{with} \\
	A^Q_n(s) &=& \argmax_{a \in A_n(s)} Q_n(s,a) ~~, \text{with} \\
	A_n(s) &=& \left\{
    \begin{array}{ll}
    \{a \in A: T_n(s,a) \leq n\} & | \text{ if not empty} \\
    \argmin_{a \in A} T_n(s,a) & | \text{ otherwise} ~,
    \end{array}\right. 
    \end{array}
\end{equation}
where $A_n(s)$, $A^Q_n(s)$, and $A^T_n(s)$ are sets of actions.
This definition of the greedy action allows the NSE to handle cyclic positive reward environments.
Moreover, it allows the selection of the best action in situations where more or less number of steps are necessary to end an episode than defined by the number of steps $n$ of a module.

To detect the greedy action in Eq.~\ref{eq:nstep_opt_action} the actions are at first limited to the set $A_n(s)$.
It is comprised of all actions resulting in trajectories that need the same or a smaller number of steps $n$ than the current module should optimize for.
If none of the actions leads to such a trajectory, then the actions that minimize the number of steps are considered.
This restriction guarantees that the NSE learns policies that end in a terminal state in cyclic environments.
For example, for module $n=1$ in the MDP of Fig.~\ref{fig:problem_mdp}, the greedy action according to the Q-value $Q_1(s_b)$ is to return to $s_b$. 
Therefore, the agent would not learn a policy ending in a terminal state if the greedy action is selected according to the Q-values.
By restricting the possible optimal actions to the set of $A_n(s)$, only actions $a_a$ and $a_b$ are allowed, because they minimize the number of steps.
As a result, the NSE learns policies that end in terminal states for modules $n=1$ and $n=2$.

The next restriction $A^Q_n(s)$ is according to the Q-values.
All actions from the set $A_n(s)$ that maximize the Q-value are considered as optimal action.
This restriction selects the trajectories resulting in the highest reward within $n$ steps.

Although, the actions $A^Q_n(s)$ maximize the reward within $n$ steps there can be situations where the resulting trajectory requires more steps or less steps.
In these situations, the Q-value does not inform about the total reward and needed time.
For example, for module $n=1$ the Q-values for going left ($a_a$) and right ($a_c$) are both zero because the Q-values of this module only look one step into the future.
Nonetheless, the final trajectory for going left result in $2$ steps and a reward of $0$, whereas for going right in $2$ steps and a reward of $1$.
In this situation it would be more desirable to go right to collect some reward.
Therefore, if several actions maximize the Q-values, then from those actions the ones resulting in the minimal number of steps $T_n$ to a terminal state are considered as defined by set $A^T_n(s)$.
Among those, the final optimal action $a^*_n(s)$ is one that has the maximum total reward $R_n$.
If several actions fulfill this, then one of them is randomly chosen.

\begin{algorithm}[t!]
\SetAlgoLined
\DontPrintSemicolon
\SetKw{KwFrom}{from}
\KwIn{\\
~~Number of modules: $M \in \N$ \\
~~Learning rate: $\alpha \in [0,1]$ \\
~~Exploration rate: $\epsilon \in [0,1]$}
\BlankLine

initialize $Q_n(s,a)$, $R_n(s,a)$, and $T_n(s,a)$ to zero 

\Repeat ({~(for each episode)}){termination}{
	initialize state $s$, and objective $f_k$\\
	
	\BlankLine
	
	\tcp{select as active module $\tilde{n}$ a module that maximizes $f_k$}
	
	$\tilde{N} \leftarrow \argmax_{n} f_k\big(R_{n}(s), T_{n}(s)\big)$
	
	$\tilde{n} \leftarrow \argmin_{n \in \tilde{N}} T_n(s)$

	\BlankLine
	
	\Repeat ({~(for each step $t$ in episode)}){$s$ is terminal-state}{

		
		\tcp{choose action $\epsilon$-Greedy}
        
		$a \leftarrow 	\begin{cases}  
						a^*_{n}(s) ~(\text{Eq. \ref{eq:nstep_opt_action}}) & | \text{ with } \Pr = 1-\epsilon \\
						\text{random action} & | \text{ with } \Pr = \epsilon					
						\end{cases}$

		\BlankLine
		
		$r$, $s' \leftarrow$ take action $a$, observe outcome

		\BlankLine

		
		\ForAll {$n \in (1, \ldots, M)$} {
		
		
		

			$Q_n(s,a) \leftarrow Q_n(s,a) + \alpha (r + Q_{n-1}(s',a^*_{n-1}(s')) - Q_n(s,a))$ 

			$R_n(s,a) \leftarrow R_n(s,a) + \alpha (r + R_{n-1}(s',a^*_{n-1}(s')) - R_n(s,a))$ 
			
			$T_n(s,a) \leftarrow T_n(s,a) + \alpha (r + T_{n-1}(s',a^*_{n-1}(s')) - T_n(s,a))$ 
			
		}
		$s \leftarrow s'$

		
		\tcp{use appropriate module for the next step}

		$\tilde{n} \leftarrow \max(1, \tilde{n}-1)$

	}
}
\caption{n-Step Ensemble (NSE)}
\label{alg:nse}
\end{algorithm}

The final policy of the NSE $\pi_{\tilde{n}}$ is given by selecting at the beginning of an episode the module that optimizes the current active objective $f_k$ according to Eq.~\ref{eq:opt_module}.
The optimal action $a^*_{\tilde{n}}$ of this module is used in the first step of the episode. 
Because each module is responsible to maximize the total reward for a certain number of steps for the next time step the optimal action of the module responsible for one step less ($\tilde{n} - 1$) is used.
This procedure is repeated until the first module $n=1$ is reached.
The policy depends therefore on the current time step $t$ during an episode:
\begin{equation*}
    \pi_{\tilde{n}}(s_t) = a^*_{n}(s_{t}) ~~\text{,with}~ n = \max(1,\tilde{n}-t) ~.
\end{equation*}
During the learning, the agent does not always chose the greedy action.
Instead an $\epsilon$-greedy action selection is used to allow exploration.

The learning of the Q-values $Q_n$, expected reward $R_n$, and number of steps $T_n$ for each module is done in a similar way to Q-learning. 
After a transition is observed all modules $n =(1,\ldots, M)$  are updated in parallel according to a Robbins-Monroe update.

\subsection{Time-dependent Q-Learning}

The IGE and NSE were compared to a classical Q-learning approach that learned for each objective an independent Q-function (Alg.~\ref{alg:ga_q}).
Its reward function $\phi$ was defined by the outcome of the active objective function $f_k$ after the agent reached a terminal state $S^G$.
For every other transition a reward of zero was given:
\begin{equation*}
	\phi_t(s_t) =  \left\lbrace \begin{array}{cl} 0 		& |~ s_t \notin S^G\\ 
											f_k(R,T)	& |~ s_t \in S^G ~,
						   \end{array} \right.
\end{equation*}
where $R$ is the collected total reward according to the reward function of the MDP and $T$ is the number of steps of the current episode.

\begin{algorithm}[t!]
\caption{Time-dependent Q-learning for general OA-MDPs}
\label{alg:ga_q}
\SetAlgoLined
\DontPrintSemicolon
\SetKw{KwFrom}{from}
\KwIn{\\
\Indp 
Learning rate: $\alpha \in [0,1]$ \\
Discount factor: $\gamma \in [0,1]$ \\
}
\BlankLine
initialize $Q(k,t,s,a)$ to zero \\
\Repeat ({~(for each episode)}){termination}{
	initialize state $s$, and goal $f_k$\\
	
	\BlankLine
	
	\tcp{start with an empty reward history}
	$h \leftarrow \emptyset$
	
	\Repeat ({~(for each step $t$ in episode)}){$s$ is terminal-state}{

		$a \leftarrow$ choose an action for $s$ derived from $Q(k,t,s,a)$ (e.g. $\epsilon$-greedy) 	

		\BlankLine
	
		\tcp{take action and save reward in history}
		
		$r$, $s'$, $isTerminal \leftarrow$ take action $a$, observe outcome
		
		$h_t \leftarrow r$
		
		\BlankLine
		
		\tcp{if a terminal state is reached, use the outcome for the}
		\tcp{objective $f_k$ as basis for the Q-function}
		
		\uIf {$isTerminal = true$} {		
		
			$\xi \leftarrow f_k(h)$ or $f_k(\sum_{i=0}^{t-1} h_i, t)$

		}
		\Else{
			$\xi \leftarrow 0$
		}

		\BlankLine
			
		$Q(k,t,s,a) \leftarrow Q(k,t,s,a) + \alpha \left(\xi + \gamma \max_{a'} Q(k,t+1,s',a') - Q(k,t,s,a)\right)$ 
		
		$s \leftarrow s'$
	}
}
\end{algorithm}

A problem with this formulation of the reward is that it does not fulfill the Markov assumption, i.e.\ that the outcome of an action depends only on the current state.
Instead, the reward $\phi$ for reaching a terminal state depends on the whole trajectory, taking into account the total reward sum $R$ and the number of steps $T$.
As a result, the MDP is partially observable for the agent because the state, i.e.\ the agent's position, does not inform about the collected return or how many steps were needed.
Therefore, Q-learning is not guaranteed to converge to the optimal policy.
To reduce this problem the current time step $t$ was used as an extra state dimension for the Q-learning agent, improving its performance.
Although the time step information improves the performance of the Q-learning agent, it does not fully resolve the partial observable nature of the problem for the agent because the collected reward sum $R$ is still missing.
Adding this information also into the state information would create a huge state space for which learning is impractical. 

\section{Experiments}

\subsection{Experimental Procedure}

The IGE, NSE and the time-dependent Q-learning agent were evaluated in the stochastic MDP illustrated in Fig.~\ref{fig:mdp}.
The agent moved with $\Pr=0.1$ in a random direction instead of the intended one.
Agents had to adapt to 9 different objectives $J = (f_1, \ldots, f_9)$ which are listed in Fig.~\ref{fig:all_results}.
The first objective $f_1$ is to receive the maximum reward in an episode.
The second $f_2$ also maximizes reward, but a punishment of $-1$ for each step is given after 3 steps.
$f_3$ gives exponentially increasing punishment for more than 3 steps.
The goal of $f_4$ is to find the shortest path to the closest terminal state. 
For $f_5$ the shortest path to a terminal state that gives at least a reward of $6.5$ is optimal.
Reaching a terminal state with less reward will result in a strong punishment.
For $f_6$ the goal is to find the highest reward with a maximum of 7 steps.
For $f_7$ the agent has only a maximum of 5 steps.
In $f_8$ the goal is to maximize average reward.
The final objective $f_9$ maximizes the average reward, but the agent has to reach at least a reward of $6.5$.
The location of the terminal states and their rewards were chosen so that they represent different solutions for the objective functions. 

For each algorithm, 100 runs were performed to measure their average learning performance.
Each run consisted of $54{,}000$ episodes that were divided in 9 phases.
In each phase, the agents had to adapt to a different objective function.
The objectives did not change during the $6000$ episodes of a phase to evaluate how long an agent needs to adapt to each objective.
The performance was measured by the outcome for the objective function $f_k(h)$ that the agent received for each episode during the learning process.

\subsection{Learning Parameters}

Learning parameters of all algorithms were manually optimized to yield a high asymptotic performance while having a high learning rate.
The learning rate parameter $\alpha$ of all algorithms was set to $\alpha=1$ in the beginning of learning to allow a faster convergence of the values.
Over the course of learning it was reduced to $\alpha=0.1$.
The IGE and NSE kept $\alpha=1$ for 500 episodes and reduced it linearly to $\alpha=0.1$ until episode 1000.
The learning rate stayed at $\alpha=0.1$ for the rest of Phase 1 and for all following phases.
The Q-learning approach needed a longer learning time to reach its asymptotic performance in each phase.
Moreover, it needed to learn a new policy for each phase.
Its learning was kept to $\alpha=1$ for 750 episodes and linearly reduced to $\alpha=0.1$ until episode 3000 for each phase.

All algorithms used the $\epsilon$-greedy action selection.
Similar to the learning rate, the exploration rate $\epsilon$ was high at the start of learning and then reduced.
The IGE's and NSE's exploration rate was $\epsilon=0.9$ for the first 500 episodes of Phase 1 and then reduced to $\epsilon=0$ until episode 1000.
It stayed at $\epsilon = 0$ for the rest of Phase 1 and all successive phases.
Q-learning used a learning rate of $\epsilon=0.9$ for the first 750 episodes.
Afterward, it was linearly reduced to $\epsilon=0$ until episode 3000 for each phase.

The IGE used a set of 45 $\gamma$-modules. 
The discount factors were chosen to have a stronger concentration for higher factors.
This allows to learn different policies for longer trajectories.
First, 14 modules were chosen according to $\gamma_{i=1:14}=\frac{i}{i+1}$. 
Then 2 modules with equal distance to each were added between each pair of the 14 modules and between the last $\gamma_{14}$ and 1.
The NSE used 20 modules.
The discount factor of Q-learning was set to $\gamma = 0.99$.

\section{Experimental Results}

\begin{figure*}[b!]
\centering

\setlength\tabcolsep{1pt}
\begin{tabular}{@{}cccc@{}}
$f_1 = R$ & 
$f_2 = \left\lbrace \begin{array}{cl} R 	& |~ T \leq 3\\ R-(T-3)& |~ T > 3 \end{array} \right.$ & 
$f_3 = \left\lbrace \begin{array}{cl} R		& |~ T \leq 3\\ R - 1.3^{(T-3)}	& |~ T > 3\end{array}\right.$ \\ 

\raisebox{-.5\height}{\includegraphics[width=0.32\linewidth]{figures/mdp125_results_learncurve_phase1}} & 
\raisebox{-.5\height}{\includegraphics[width=0.32\linewidth]{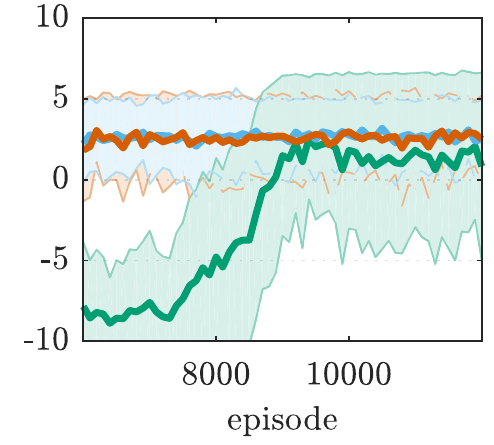}} & 
\raisebox{-.5\height}{\includegraphics[width=0.32\linewidth]{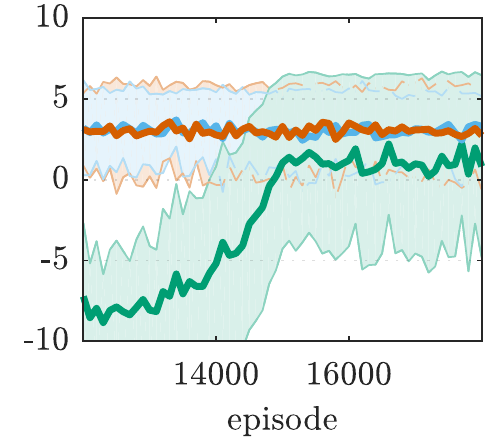}} \\ [2.5cm]

$f_4 = -T$ &
$f_5 = \left\lbrace \begin{array}{cl}-10& |~ R \leq 6.5\\ -T	& |~ R > 6.5\end{array}\right.$ & 
$f_6 = \left\lbrace \begin{array}{cl}R   & |~ T \leq 7\\ -10& |~ T > 7\end{array}\right.$ \\
 
\raisebox{-.5\height}{\includegraphics[width=0.32\linewidth]{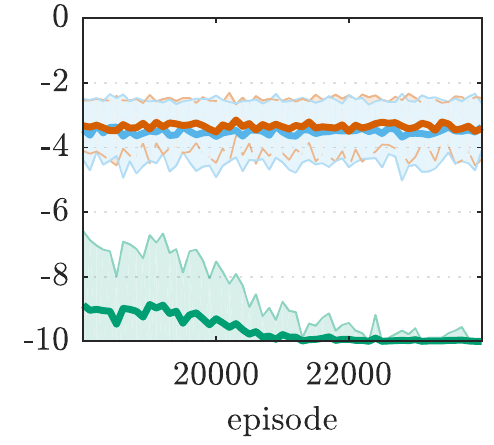}} &
\raisebox{-.5\height}{\includegraphics[width=0.32\linewidth]{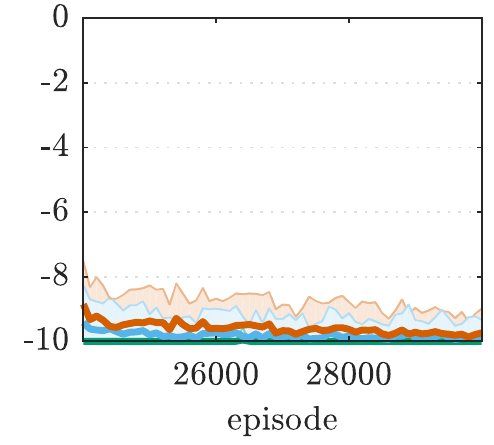}} & 
\raisebox{-.5\height}{\includegraphics[width=0.32\linewidth]{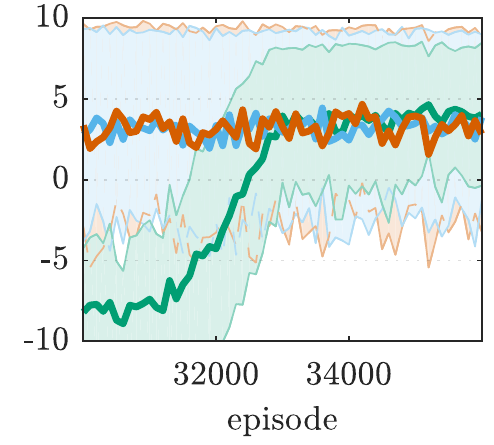}} \\ [2.5cm] 

$f_7 = \left\lbrace \begin{array}{cl}R   & |~ T \leq 5\\ -10& |~ T > 5 \end{array} \right.$ & 
$f_8 = \frac{R}{T}$ &
$f_9 = \left\lbrace \begin{array}{cl}\frac{R}{T}   & |~ R \geq 6.5\\ -1& |~ R < 6.5 \end{array} \right.$ \\

\raisebox{-.5\height}{\includegraphics[width=0.32\linewidth]{figures/mdp125_results_learncurve_phase7}} & 
\raisebox{-.5\height}{\includegraphics[width=0.32\linewidth]{figures/mdp125_results_learncurve_phase8}} & 
\raisebox{-.5\height}{  \begin{overpic}[width=0.32\linewidth]{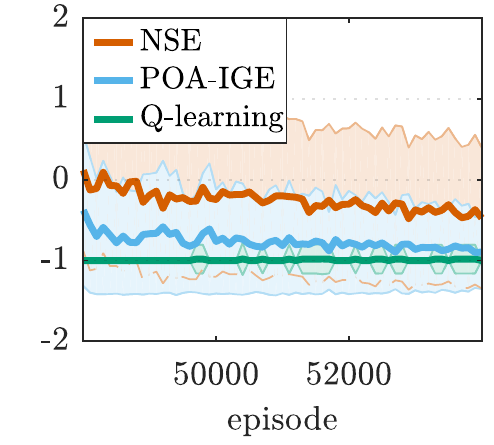}
                            \put(16.5,60.5){\includegraphics[width=0.135\linewidth]{figures/legend.png}}  
                        \end{overpic}
                    } \\
  
\end{tabular}

 \caption[Results for objective-adaptive algorithms in general tasks]
 {
	The IGE and NSE immediately adapted to new objective functions compared to the time-dependent Q-learning approach in the stochastic task of Fig.~\ref{fig:mdp}.
	Performance was measured by the outcome of the objective function $f_k(R,T)$ per episode, where $R=\sum_{t=0}^{T-1} r_t$ is the reward sum of the agent's trajectory and $T$ is its length.
	Each of the 9 phases has a different objective function.
	The plots show the mean and standard deviation over $100$ runs per algorithm.
	The minimal reward per episode was limited to $-10$ to make the plots more readable, because some goal formulations can result in a large negative reward during explorations.
 } 
 \label{fig:all_results}
\end{figure*}

The results show that the IGE and NSE performed better compared to Q-learning in terms of adaptation to new objectives, asymptotic performance and learning speed (Fig.~\ref{fig:all_results}).
After the initial learning phase with objective $f_1$ the IGE and NSE were able to adapt immediately to new objectives whereas Q-learning needed to learn new policies for those phases.

Moreover, the IGE and NSE outperformed Q-learning in their asymptotic performance in 5 of the 9 objectives ($f_1$, $f_2$, $f_3$, $f_4$, $f_9$).
All algorithms had a similar final performance for 3 objectives ($f_5$, $f_6$, $f_8$).
Q-learning could slightly outperform IGE for objective $f_7$, but not the NSE.
Comparing the IGE and the NSE, the NSE had a slightly better final performance for $f_7$ and a stronger performance for $f_9$.
The low performance of Q-learning for objective $f_4$ was the result of the negative outcome values that this objective gives. 
Q-values were initialized to 0.
Because all outcomes for this objective are negative, the agent is doing an optimistic exploration \citep{osband2017optimistic}.
It explores every possible state action pair for every possible time step, because it has to learn that their initial Q-values of 0 are not optimal.
As a result, Q-learning would need more episodes to learn a good policy for this objective.

As for the learning rate, the IGE and NSE outperformed Q-learning which is visible in the first phase.
Q-learning needed at least 3000 episodes to reach its final asymptotic performance for each phase.
The IGE and NSE needed less than 1000 episodes to reach their asymptotic performance in the first phase.
Q-learning needed longer due to the extra time step information in its state space. 
Thus, more exploration was necessary to learn the optimal policies.

\end{document}

%% file: math_commands.tex
\usepackage{amsmath,amsfonts,bm}

\def\eqref#1{equation~\ref{#1}}

\def\1{\bm{1}}

\DeclareMathAlphabet{\mathsfit}{\encodingdefault}{\sfdefault}{m}{sl}
\SetMathAlphabet{\mathsfit}{bold}{\encodingdefault}{\sfdefault}{bx}{n}

\newcommand{\E}{\mathbb{E}}

\newcommand{\R}{\mathbb{R}}

\DeclareMathOperator*{\argmax}{arg\,max}
\DeclareMathOperator*{\argmin}{arg\,min}